\documentclass[journal]{IEEEtran}
\usepackage{graphicx}
\usepackage{amsmath}
\usepackage[caption=false,font=footnotesize]{subfig}
\usepackage{amsmath}

\begin{document}
	
	\title{Added value of morphological features to breast lesion diagnosis in ultrasound}
	
	\author{Micha\l{}~Byra,
		Katarzyna Dobruch-Sobczak,
		Hanna Piotrzkowska-Wr\'{o}blewska
		and~Andrzej~Nowicki%
		\thanks{M. Byra, K. Dobruch-Sobczak, H. Piotrzkowska-Wr\'{o}blewska and A. Nowicki are with the Department of Ultrasound, Institute of Fundamental Technological Research, Polish Academy of Sciences, Pawi\'{n}skiego 5B, 02-106, Warsaw, Poland.}
		\thanks{Corresponding author: M. Byra, mbyra@ippt.pan.pl}}

	\maketitle
	
	\begin{abstract}
		
		Ultrasound imaging plays an important role in breast lesion differentiation. However, diagnostic accuracy depends on ultrasonographer experience. Various computer aided diagnosis systems has been developed to improve breast cancer detection and reduce the number of unnecessary biopsies. In this study, our aim was to improve breast lesion classification based on the BI-RADS (Breast Imaging – Reporting and Data System). This was accomplished by combining the BI-RADS with morphological features which assess lesion boundary. A dataset of 214 lesion images was used for analysis. 30 morphological features were extracted and feature selection scheme was
		applied to find features which improve the BI-RADS classification performance. Additionally, the best performing
		morphological feature subset was indicated. We obtained a better classification by combining the BI-RADS with six morphological features. These features were the extent, overlap ratio,
		NRL entropy, circularity, elliptic-normalized circumference and the normalized residual value. The area
		under the receiver operating curve calculated with the use of the combined classifier was 0.986. The
		best performing morphological feature subset contained six features: the DWR, NRL entropy,
		normalized residual value, overlap ratio, extent and the morphological closing ratio. For this set, the area
		under the curve was 0.901. The combination of the radiologist's experience related to the BI-RADS and the morphological features leads to a more effective breast lesion classification.
	
	\end{abstract}

	
	\IEEEpeerreviewmaketitle
	
	\section{Introduction}
	
	\IEEEPARstart{A}{ccording} to the World Health Organization, breast cancer is one of the most frequent cancer diseases in the world \cite{Stewart2014}. To provide more effective treatment and reduce the death rate, early
	detection of breast cancer must be carried out. Despite the fact that different diagnostic tools can be used
	to detect breast cancer, there is a growing interest in use of ultrasound imaging. Ultrasound is known
	for being non-invasive, relatively non-expensive and broadly accessible. As opposed to mammography
	which is not sensitive in the case of women with dense breast. As it was demonstrated in several studies \cite{Kolb2002, Stavros1995, Zhi2007}, ultrasound can be successfully used for breast cancer detection. However, diagnosis conducted by
	means of ultrasound imaging requires experienced radiologists who know how to efficiently operate ultrasound
	scanner and possess knowledge of breast cancer heterogeneity and its complex characteristic
	features appearing on ultrasound images. Therefore, many unnecessary biopsies are performed.
	
	To
	standardize the reporting process and diagnosis, American College of Radiology developed a quality
	control system called BI-RADS (Breast Imaging – Reporting and Data System) which is now widely
	accepted and used by physicians \cite{Bott2014}. After the interpretation of the lesion ultrasound image, a specific BIRADS
	category is assigned which reflects the risk of malignancy. However, this assessment still
	depends on the ultrasonographer’s experience and his ability to interpret the ultrasound image correctly.
	Therefore, computer-aided diagnosis (CAD) systems are investigated to improve the breast lesion
	classification and support physicians, especially the inexperienced ones.
	
	The main goal of CAD is to develop a computer program which would be able to differentiate
	breast lesions based on ultrasound images analysis \cite{Cheng2010}. The common approach is to extract features
	from the image which contains the lesion and then develop a classifier using machine learning methods.
	Well-chosen features are the most important part of every CAD system. So far, various sets of features
	were proposed in the literature for the breast lesion classification. Those features are primarily divided into two categories, namely the
	texture and the morphological features. Nowadays, morphological features are considered to be the most
	effective in breast cancer classification \cite{GomezFlores2015}, though good performance was also reported for other features.
	Morphological features assess lesion contour. Generally, more irregular contour is expected in the case
	of malignant lesions.
	
	Morphological features have some indisputable advantages, especially in comparison with
	texture features. They are less affected by image processing algorithms used for B-mode image
	reconstruction. Many CAD systems were developed based on B-mode images acquired with a single
	ultrasound machine. However, usually little is known of the image reconstruction algorithms
	implemented in the scanner. Most of the ultrasound image enhancing algorithms intensively process
	texture \cite{ContrerasOrtiz2012} which may have negative impact on the classification performed with texture features. On the
	contrary, edge preservation and emphasis is one of the main goals of image processing algorithms what
	places morphological features in a far better position than texture features. Texture features depend on
	operator and particular machine settings, e.g. focal depth positioning \cite{Garra1993}. Estimation of these features can
	be affected additionally by the presence of calcifications or necrosis within the lesion \cite{Byra2016, Larrue2014917}.
	
	The aim of this work is to combine the BI-RADS with morphological features to
	improve the classification. CAD papers usually don’t utilize BI-RADS categories which were assigned
	by the radiologist. This practice is understandable. While the process of features extraction is well
	defined mathematically, the assignation of a specific BI-RADS category depends on radiologist’s
	experience. This subjective assessment may affect the CAD system performance and make the
	comparison with other CAD systems problematic. On the other hand, it is of great importance to
	investigate whether a CAD system can support physicians. The widely used BI-RADS  has its
	limitations which might be overcome with the CAD. Here, we investigate whether the morphological
	features can improve the BI-RADS classification performance or if they are rather redundant. First, the
	radiologist assigned the BI-RADS category to each lesion. Next, to improve the classification
	morphological features were chosen and combined with the BI-RADS. The performance of the
	combined classifier was compared with the BI-RADS and the best performing morphological
	feature subset.
	
	This paper is organized in a following way. In the first section, the breast lesion database and
	the acquisition procedure are described. Next, we give a list of investigated morphological features
	including the papers in which they were originally proposed or later used. We present the scheme for
	feature selection. Then, we present the best performing feature subset. The same scheme is used to find
	which morphological features may improve the performance of BI-RADS. Finally, we discuss results
	and present conclusions.
	
	\section{Materials and Methods}	
	
	\subsection{Dataset and preprocessing}		
	
	The database consists of 214 images of 107 solid lesions, 75 of which are benign and 32
	malignant. For each lesion, two perpendicular scan planes were acquired during routine breast diagnostic
	procedures. Ultrasonix scanner (Ultrasonix Inc., Canada) equipped with a linear array probe L14-5/38
	was used to collect the data. The focal depth was positioned at the center of each lesion. The imaging
	frequency was set to 7.5 MHz. Each lesion was biopsy proven. First, however, a specific BI-RADS
	category was assigned by the radiologist with 17 years’ experience in ultrasonic diagnosis of breast
	lesions. The BI-RADS has 7 categories which reflect the likelihood of malignancy \cite{Bott2014, W.2012}:
	
	\begin{table}[b]
		\centering
		\caption{Breast lesions BI-RADS categories.}	
		\begin{tabular}{|c|c|c|c|c|c|}
			\hline
			BI-RADS & 3 & 4a & 4b & 4c & 5 \\    \hline
			Benign & 41 & 19 & 14 & 0 & 1 \\    \hline
			Malignant & 0 & 1 & 5 & 6 & 20 \\    \hline
		\end{tabular}
	\end{table}	
	
	\begin{itemize}
		\item 0: incomplete
		\item 1: negative
		\item 2: benign 
		\item 3: probably benign
		\item 4: suspicious (4a - low suspicious, 4b - intermediate suspictious, 4c - moderate suspicious)
		\item 5: probably malignant
		\item 6: malignant
	\end{itemize}

	\begin{table}[b!]
		\centering
		\caption{Implemented morphological features.}	
		\begin{tabular}{|c|l|}
			\hline
			Number & Feature \\    \hline
			1 & Angular characteristics \cite{Shen2007} \\    \hline
			2 & Area ratio \cite{Alvarenga2012, Alvarenga2010, Chou2001}\\    \hline
			3 & Aspect ratio \cite{Chang2005} \\    \hline
			4 & Branch pattern \cite{Joo2004} \\    \hline
			5 & Circularity \cite{Chang2005, Alvarenga2010, Chou2001, Drukker2004a, Kim2002, Drukker2005}\\    \hline
			6 & Contour roughness \cite{Alvarenga2012, Alvarenga2010, Chou2001} \\    \hline
			7 & Convexity  \cite{Chang2005}\\    \hline
			8 & DWR \cite{Drukker2004a, Drukker2002, Drukker2005, Horsch2002a, Chen2003, Chen2004} \\    \hline
			9 & Ellipsoidal shape \cite{Joo2004, Su2011}\\    \hline
			10 & Elliptic-normalized circumference \cite{Chen2003}\\    \hline
			11 & Elliptic-normalized skeleton \cite{Chen2003}\\    \hline
			12 & Extent \cite{Chang2005} \\    \hline
			13 & Lesion size \cite{Chen2003}\\    \hline
			14 & Lobulation index \cite{Chen2003}\\    \hline
			15 & Long to short axis ratio \cite{Chen2003}\\    \hline
			16 & Morphological closing ratio \cite{Alvarenga2010}\\    \hline
			17 & Normalized residual value \cite{Alvarenga2010}\\    \hline
			18 & NRL entropy \cite{Chou2001}\\    \hline
			19 & NRL mean \cite{Chou2001}\\    \hline
			20 & NRL standard deviation \cite{Alvarenga2012, Alvarenga2010, Chou2001}\\    \hline
			21 & NRL zero-crossing \cite{Chou2001}\\    \hline
			22 & Number of lobulations \cite{Joo2004, Su2011}\\    \hline
			23 & Number of substantial protuberances and depressions \cite{Chen2003}\\    \hline
			24 & Orientation \cite{Shen2007}\\    \hline
			25 & Overlap ratio \cite{Alvarenga2010}\\    \hline
			26 & Roundness \cite{Chang2005}\\    \hline
			27 & Shape class \cite{Shen2007, Minavathi2012}\\    \hline
			28 & Solidity \cite{Chang2005} \\    \hline
			29 & Spiculation \cite{Joo2004, Su2011}\\    \hline
			30 & Undulation characteristics \cite{Shen2007} \\    \hline															
		\end{tabular}
	\end{table}			
	
	\noindent	In our study, BI-RADS categories of the lesions varied from 3 to 5 as it is depicted in Table 1.
	
	Initial contour was indicated by the physician and subsequently improved with a computer
	algorithm \cite{Chan2001}, see Fig. 1. All calculations were performed in Matlab (The MathWorks Inc.)

	\subsection{Features}
	
	\begin{figure*}[t!]
		\centering
		\includegraphics[width=6in]{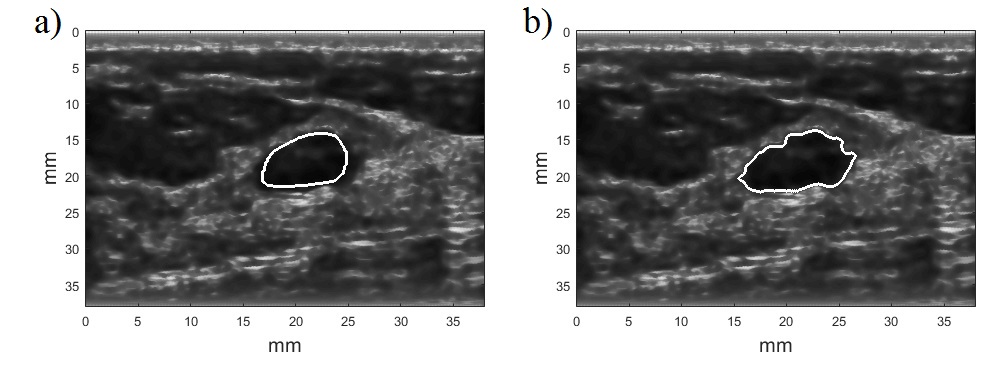}%
		\caption{The contour drawn manually a) and the contour improved with the computer algorithm b).}
		\label{fig1}
		\hfil
	\end{figure*}
	
	We implemented 30 morphological features, which are listed in Table 2. For each feature, papers
	in which it was developed or used are mentioned. Here, we describe the most popular features. However,
	for an in-depth analysis, reader is referred to the cited works.
	
	Depth to width ratio (DWR) is perhaps the most frequently used morphological feature. It is
	easy to calculate and was reported to be effective in many papers. The depth and the width are the
	dimensions of the minimal circumscribed rectangle which contains the lesion. Some papers use the
	inversion of the DWR \cite{Chang2005}, but here it will not be treated as a separate feature. A feature similar to the
	DWR is the long to short axis ratio of the ellipse inscribed in the lesion contour.
	
	The normalized radial length (NRL) is defined by the following equation:
	
	\begin{equation}
	d_n(i) = \frac{d(i)}{\text{max}[d(i)]}
	\end{equation}
	
	\noindent where $d(i)$ is the distance from the lesion’s center of mass to $i$-th point on its perimeter. The NRL is
	used to obtain features which measure various properties of the contour, see positions 2, 6, 18-21 in
	Table 2.
	
	Some features, namely 7, 17, 23, 25 and 28, are based on the convex hull of the lesion. These
	features were introduced to measure the level of spiculations which is reflected by the protuberance of
	the contour.
	
	To take into account that we possess two scans of the same lesion, each feature was calculated
	for both scan plane and results were averaged.
			
	\subsection{Classification and evaluation}
	
	We explore the performance of morphological features in multiple ways. As it was shown in
	several papers \cite{GomezFlores2015, daoud2016fusion}, there is no single feature which would alone outperform the others, therefore features
	must be combined in order to improve the classification. However, for a large number of features it is
	problematic to perform an exhaustive search for the best performing subset due to a large number of all
	potential combinations. Here, we used a two-step approach to find the best subset for classification. The
	goal was to maximize the area under the curve (AUC) of the receiver operating characteristics (ROC).
	In the first step, the best performing feature (the highest value of AUC) was chosen. Next, the first feature was combined with the remaining features in order to select the best performing subset of two
	features. This forward selection procedure was repeated until the feature pool was empty. We decided
	to maximize the AUC because this quantity is after all usually used for classification performance
	assessment. To evaluate AUC we applied the leave-one-out cross-validation. Logistic regression was
	used to perform classification. Before the training, features were standardized. To address the class
	imbalance, during cost function minimization, sampling weights were inversely proportional to class
	frequencies in training set. For each test sample the probability of malignancy was calculated. The
	AUC standard deviation was calculated with the bootstrap method. Next, we performed the second step
	of the feature selection procedure which was the backward selection. Supposedly, different feature
	subsets may have similar AUC values. In this case, we selected the best performing subset, the one with
	the highest AUC value. Next, the ANOVA analysis along with the Tukey test were used to find a subset
	with a smaller number of features and likely the same mean AUC value as the best-performing subset
	at 95\% confidence level.
	
	With the above feature selection methodology, first the best morphological feature subset was
	selected. Next, the BI-RADS was combined with morphological features. However, to use the
	BI-RADS category as a feature, some kind of transformation must be performed. In our case the BI-RADS was
	treated as a discrete feature which can be coded with integers. We used the following scheme: integer 1
	stands for BI-RADS category 1, 2 for 2, 3 for 3, 4 for 4a, 5 for 4b, 6 for 4c, 7 for 5 and 8 for 6,
	respectively.
		
	\section{Results}
	
	\begin{figure*}[]
		\centering
		\includegraphics[width=5.5in]{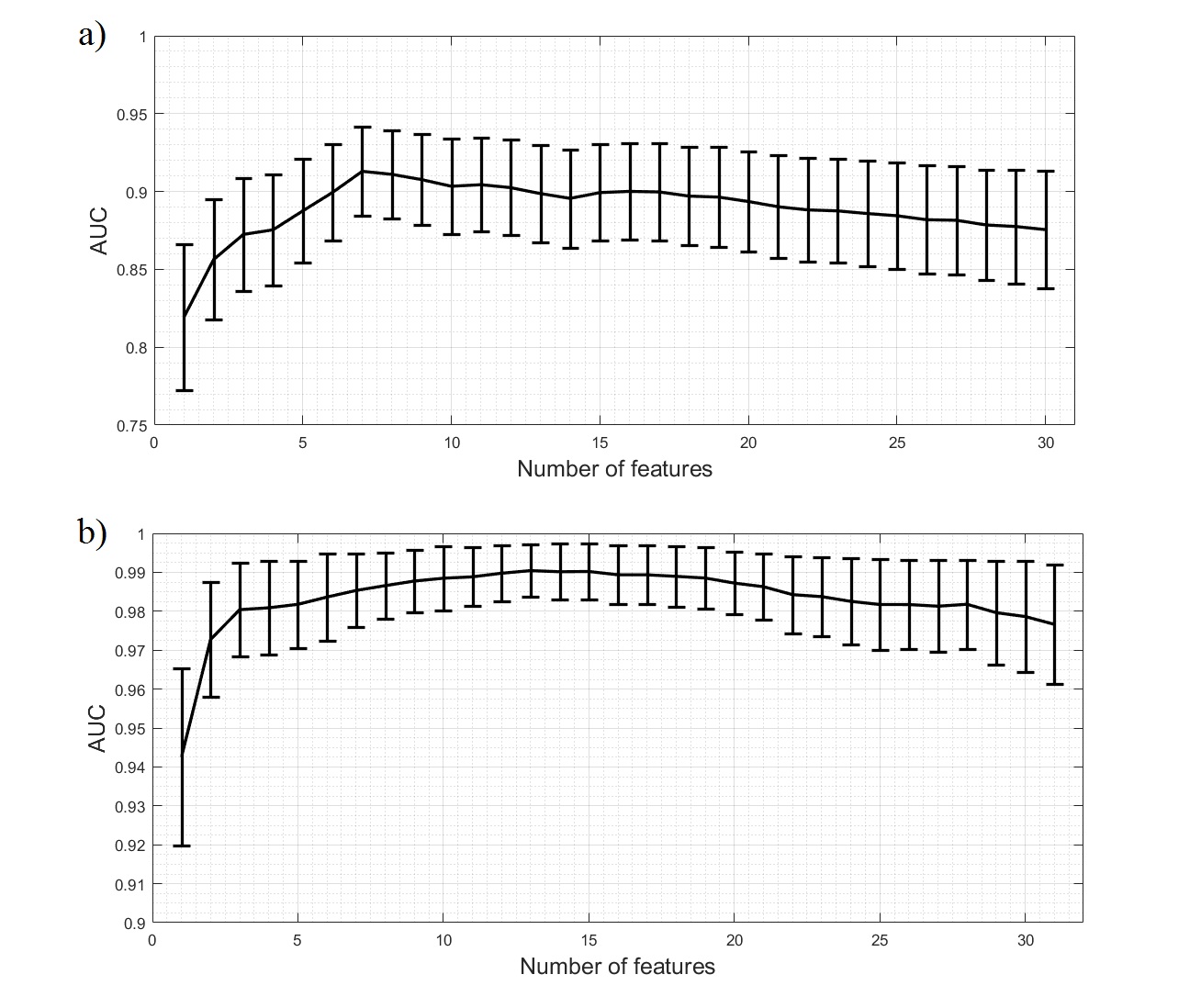}%
		\caption{AUC estimation ($\pm$ standard deviation) for features ranked with the proposed criterion: a) morphological features and b) morphological features with the BI-RADS.}
		\label{fig2}
		\hfil
	\end{figure*}		
	
	The feature selection procedure is depicted in Fig. 2. The highest value of AUC in the case of
	morphological features, see Fig.2 a), was obtained for a set of seven features, however the ANOVA
	analysis showed that there is no significant difference between this set and a subset containing six
	features. Therefore, the smaller subset was selected as the best performing. It consisted of six features
	which were the DWR, NRL entropy, normalized residual value, overlap ratio, extent and the
	morphological closing ratio. Similar analysis was performed to combine the BI-RADS with
	morphological features. The highest AUC was obtained for a 13 feature subset (including BI-RADS) which was then reduced to seven. Features which, when added to the BI-RADS, improved the
	classification most, were the extent, overlap ratio, NRL entropy, circularity, elliptic-normalized
	circumference and the normalized residual value.

	Main results are depicted in Table 3. The use of the best performing morphological feature
	subset and the BI-RADS resulted in the AUC values of 0.901 and 0.944, respectively. The classification
	was improved when the BI-RADS and morphological features were combined. With six features added
	to the BI-RADS, the AUC value increased to 0.986. Fig. 3 shows the ROC curves obtained for the
	developed classifiers. The optimal sensitivity, specificity and accuracy of each classifier was determined
	by means of the ROC curve for the point which was the closest to (0, 1) \cite{Fawcett2006}.
	\begin{table*}[]
		\centering
		\caption{The summary. Numeration of morphological features according to Table 2.}	
		\begin{tabular}{|c|c|c|c|}
			\hline
			Features & Sensitivity [\%] & Specificity [\%] & Accuracy [\%]  \\    \hline
			Morphological, optimal cut-off & 75.0 & 88.0 & 84.1  \\    \hline
			Combination, optimal cut-off & 96.8 & 94.7 & 95.3  \\    \hline
			Morphological, customized cut-off & 100 & 58.7 & 71.0 \\    \hline
			Combination, customized cut-off & 100 & 74.7 & 82.2  \\    \hline
			BI-RADS cat. 3 cut-off & 100 & 54.7 & 68.2 \\    \hline
		\end{tabular}
	\end{table*}	
	
	In the case of the breast lesion classification, it is important to have as high sensitivity as possible
	to detect all malignant lesions. According to Table 1 in the case of the BI-RADS, 100\%
	sensitivity could only be obtained when lesions with the BI-RADS category higher than 3 were classified
	as malignant. The corresponding specificity was 54.7\%. Taking this into consideration, the thresholds of
	the classifiers were customized based on ROC curves to ensure 100\% sensitivity and the corresponding
	accuracies and specificities were calculated. Results are depicted in Table 3. The specificity of the best
	performing morphological features subset was 58.7\%, similarly to the BI-RADS. Moreover, with
	the combined classifier it was possible to obtain 100\% sensitivity and specificity of 74.7\%. Table 4 shows
	how many biopsies could be avoided in the case of benign lesions by using various classifiers with cut-offs
	ensuring 100\% sensitivity.
		
	\section{discussion}
	
	With the use of morphological features it was possible to differentiate between malignant and
	benign breast lesions. The best performing feature subset achieved AUC value of 0.901. However, this
	result was worse than in the case of the BI-RADS for which AUC value was 0.944. The classifier based
	on the morphological features could not outperform the radiologist who assigned the BI-RADS categories.
	It must be emphasized that the BI-RADS depends on the physician’s experience and for a novice
	radiologist the BI-RADS performance could be lower than the performance of the CAD system.
	However, both AUC values, obtained for the BI-RADS and for the morphological features, should be
	considered satisfactory. 
	
	\indent The best morphological subset consisted of six features which were the DWR, NRL entropy, 	normalized residual value, overlap ratio, extent and the morphological closing ratio. The DWR was
	reported as the best feature for breast lesion classification in several papers \cite{Drukker2004a, Drukker2002, Drukker2005, Horsch2002a, Chen2003, Chen2004}. NRL entropy is higher
	for lesions which have irregular contour. The normalized residual value was indicated as the best feature
	for breast lesion classification in \cite{Alvarenga2010}. This feature is based on the difference between the lesion convex
	hull area and the regular area which was normalized by convex hull perimeter. The overlap
	ratio is the ratio of the convex hull area and the lesion area. It measures contour irregularity and was reported
	as one of the best features in original work \cite{Alvarenga2010}. The extent is the ratio of lesion area to the smallest rectangle
	inscribed in it. The morphological closing ration tends to be greater for lesions which have irregular
	contour \cite{Alvarenga2010}.
	\begin{figure}[]
		\centering
		\includegraphics[width=.5\textwidth]{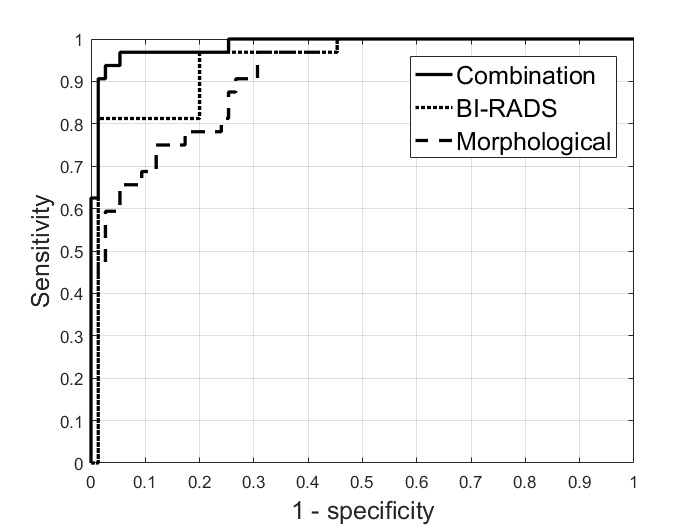}%
		\caption{ROC curves for the BI-RADS, morphological features and the combination.}
		\label{fig3}
		\hfil
	\end{figure}	
	\begin{table}
		\centering
		\caption{Number of benign lesions and biopsies that could be possible avoided in the case of each classifier.}	
		\begin{tabular}{|c|c|c|c|c|c|c|}
			\hline
			BI-RADS & 3 & 4a & 4b & 4c & 5 & Sum \\    \hline
			Nr of benign lesions & 41 & 19 & 14 & 0 & 1 & 75 \\    \hline
			Combined classifier & 0 & 1 & 5 & 6 & 20 & 56 \\    \hline
			Best morphological & 41 & 19 & 14 & 0 & 1 & 44 \\    \hline
			BI-RADS cat. 3 cut-off & 0 & 1 & 5 & 6 & 20 & 41 \\    \hline			
		\end{tabular}
	\end{table}		
	
	According to the survey investigating the performance of various features, the best
	morphological features for the breast lesion classification are the elliptic-normalized skeleton, lesion
	orientation, the number of substantial protuberances and depressions, DWR and the overlap ratio \cite{GomezFlores2015}. For
	this set, the reported AUC value was 0.94. In our case, the use of the proposed features lead to the AUC
	value of 0.871. This difference may be due to the dataset. However, this particular performance should
	also be perceived as good. The main doubt lies in the choice of the lesion orientation. This feature
	measures the angle of major axis of above lesion best-fit ellipse and is extremely operator dependent
	since the motion of the imaging probe may easily change it. In our study, the calculated AUC value for
	the orientation was 0.532. Similar result was reported in the original study \cite{Shen2007} where this feature had
	negligible impact on the classification and was the first one to be removed from the feature pool when
	applying backward remove feature selection method.
	
	A great advantage of morphological features is revealed when they are combined with the BIRADS.
	It was possible to increase the AUC value to 0.986 by adding six features, namely the extent,
	overlap ratio, NRL entropy, circularity, elliptic-normalized circumference and the normalized residual
	value. The circularity is the ratio of a lesion squared perimeter and the lesion area \cite{Alvarenga2012}. The elliptic normalized
	circumference quantify the anfractuosity of a lesion contour \cite{Chen2003}.
	
	One of the main goals of breast lesion classification is to have 100\% sensitivity and as high
	specificity as possible to indicate benign lesions and reduce the number of unnecessary biopsies. In
	our study, as all lesions were biopsy proven, therefore their evaluation must be considered problematic
	for the radiologist. The first advantage of the combined classifier is that it can be used directly to support
	the radiologist in the process of decision making. First, the radiologist assigns a specific BI-RADS
	category to the lesion, then the combined classifier containing the morphological features is used to
	indicate the level of malignancy. The decision of the radiologist is improved be means of morphological
	features. The morphological features are used to separate malignant and benign lesions that were
	assigned the same BI-RADS category. As it is shown in Fig. 3, 100\% sensitivity was obtained with high
	specificity of 74.7\%, much higher than in the case of the BI-RADS alone. According to Table 4, with
	the combined classifier it would be possible to correctly classify all benign lesions with the BI-RADS
	category 3 as in the case of the BI-RADS alone. However, in addition few examples of benign lesions
	with higher BI-RADS categories would be correctly classified. With the combined classifier, it would
	possible to avoid 56 biopsies. However, it must be stressed that there were no malignant lesions in the
	dataset with BI-RADS category 3, the classification performed by the radiologist was already at high
	level. Moreover, use of the best morphological feature subset provides higher specificity at 100\%
	sensitivity than the BI-RADS as it is shown in Table 3, although its AUC value is lower. However in
	this case it was not be possible to correctly classify all lesions with the BI-RADS category 3.
	
	The main disadvantage of the combined classifier is that it was developed based on the
	experience of a particular physician who assigned the BI-RADS categories. Supposedly, several issues
	might occur. First, although the developed CAD system can serve as support for a particular radiologist,
	it might, however, not work when used by another radiologist. For example, a less experienced physician
	can have a worse performance, which translates to a different BI-RADS ROC curve and therefore affects
	the performance of the combined classifier. Next, the choice of morphological features chosen to
	improve BI-RADS may depend on the radiologist’s experience. For example, features developed to
	assess spiculations can be selected if the radiologist does not evaluate spiculations successfully.
	In all these cases, the system would require separate training to support a particular radiologist. It would
	be interesting to utilize the combined system in the radiologists' training. Hypothetically, after the
	assignation of the BI-RADS categories to an exemplary dataset, the feature selection can be used to
	indicate which features improves the diagnosis in this particular case. For example, a “novice”,
	inexperienced radiologist (or even an experienced one) can be told to pay more attention to spiculations.
	Moreover, radiologists perceive image features differently and the tumor assessment is usually descriptive.
	Numerical values reflecting the level of spiculation (or other contour characteristic) quantitatively could
	be helpful by themselves, even without a CAD system. It could enable a more objective lesion
	description. The majority of morphological features is easy to illustrate on the image which may help
	the radiologist to analyze the lesion.			
	
	\section{conclusion}
	
	In this work, we investigated the usefulness of morphological features for the breast lesions
	differentiation. The main goal was to find features that can improve the BI-RADS. This was successfully
	accomplished by incorporating six morphological features. The use of the developed combined classifier
	leads to 100\% sensitivity and high specificity of 74.7\%. It can be used to reduce the number of
	unnecessary biopsies. The combined classifier depends on the experience of a particular radiologist,
	however, the presented in this work approach can be used to train a classifier for a different radiologist.
	Besides, other features, for instance texture features can be incorporated in the future to potentially
	improve the classification further. The developed CAD system can also be used in the radiologists'
	training. After the classifier development phase, the radiologist can be informed which features improve
	his diagnosis accuracy. This enables the radiologist to widen his knowledge of lesion appearance in ultrasound image. With the help of artificial intelligence, the radiologist can hypothetically improve his
	classification performance. 
	
	In our study we obtained good classification performance with
	particular morphological features, even without the BI-RADS system. However, in comparison with the
	survey paper \cite{GomezFlores2015}, we determined a slightly different best performing feature subset.
	
	\section*{Acknowledgments:}
	
	This work was supported by the National Science Center Grant Number UMO-	2014/13/B/ST7/01271.
	
	\section*{Conflict of interest statement}
	
	None.
	
	\bibliographystyle{ieeetran}
	\bibliography{biblio}

\begin{thebibliography}{10}
\providecommand{\url}[1]{#1}
\csname url@samestyle\endcsname
\providecommand{\newblock}{\relax}
\providecommand{\bibinfo}[2]{#2}
\providecommand{\BIBentrySTDinterwordspacing}{\spaceskip=0pt\relax}
\providecommand{\BIBentryALTinterwordstretchfactor}{4}
\providecommand{\BIBentryALTinterwordspacing}{\spaceskip=\fontdimen2\font plus
\BIBentryALTinterwordstretchfactor\fontdimen3\font minus
  \fontdimen4\font\relax}
\providecommand{\BIBforeignlanguage}[2]{{%
\expandafter\ifx\csname l@#1\endcsname\relax
\typeout{** WARNING: IEEEtran.bst: No hyphenation pattern has been}%
\typeout{** loaded for the language `#1'. Using the pattern for}%
\typeout{** the default language instead.}%
\else
\language=\csname l@#1\endcsname
\fi
#2}}
\providecommand{\BIBdecl}{\relax}
\BIBdecl

\bibitem{Stewart2014}
B.~W. Stewart and C.~P. Wild, \emph{{World Cancer Report 2014}}, 2014.

\bibitem{Kolb2002}
\BIBentryALTinterwordspacing
T.~M. Kolb, J.~Lichy, and J.~H. Newhouse, ``{Comparison of the performance of
  screening mammography, physical examination, and breast US and evaluation of
  factors that influence them: an analysis of 27,825 patient evaluations.}''
  \emph{Radiology}, vol. 225, no.~1, pp. 165--75, 2002. [Online]. Available:
  \url{http://www.ncbi.nlm.nih.gov/pubmed/12355001}
\BIBentrySTDinterwordspacing

\bibitem{Stavros1995}
A.~T. Stavros, D.~Thickman, C.~L. Rapp, M.~A. Dennis, S.~H. Parker, and G.~A.
  Sisney, ``{Solid breast nodules: use of sonography to distinguish between
  benign and malignant lesions.}'' \emph{Radiology}, vol. 196, no.~1, pp.
  123--134, 1995.

\bibitem{Zhi2007}
H.~Zhi, B.~Ou, B.-M. Luo, X.~Feng, Y.-L. Wen, and H.-Y. Yang, ``{Comparison of
  ultrasound elastography, mammography, and sonography in the diagnosis of
  solid breast lesions.}'' \emph{Journal of ultrasound in medicine : official
  journal of the American Institute of Ultrasound in Medicine}, vol.~26, no.~6,
  pp. 807--815, 2007.

\bibitem{Bott2014}
R.~Bott, \emph{{ACR BI-RADS Atlas}}, 2014, no.~1.

\bibitem{Cheng2010}
H.~D. Cheng, J.~Shan, W.~Ju, Y.~Guo, and L.~Zhang, ``{Automated breast cancer
  detection and classification using ultrasound images: A survey},''
  \emph{Pattern Recognition}, vol.~43, no.~1, pp. 299--317, 2010.

\bibitem{GomezFlores2015}
W.~{Gomez Flores}, W.~C. D.~A. Pereira, and A.~F.~C. Infantosi, ``{Improving
  classification performance of breast lesions on ultrasonography},''
  \emph{Pattern Recognition}, vol.~48, no.~4, pp. 1121--1132, 2015.

\bibitem{ContrerasOrtiz2012}
S.~H. {Contreras Ortiz}, T.~Chiu, and M.~D. Fox, ``{Ultrasound image
  enhancement: A review},'' pp. 419--428, 2012.

\bibitem{Garra1993}
B.~S. Garra, B.~H. Krasner, S.~C. Horii, S.~Ascher, S.~K. Mun, and R.~K. Zeman,
  ``{Improving the distinction between benign and malignant breast lesions: the
  value of sonographic texture analysis.}'' \emph{Ultrasonic imaging}, vol.~15,
  no.~4, pp. 267--285, 1993.

\bibitem{Byra2016}
M.~Byra, A.~Nowicki, H.~Wr{\'{o}}blewska-Piotrzkowska, and K.~Dobruch-Sobczak,
  ``{Classification of breast lesions using segmented quantitative ultrasound
  maps of homodyned K distribution parameters},'' \emph{Medical Physics},
  vol.~43, no.~10, pp. 5561--5569, oct 2016.

\bibitem{Larrue2014917}
\BIBentryALTinterwordspacing
A.~Larrue and J.~A. Noble, ``{Modeling of Errors in Nakagami Imaging:
  Illustration on Breast Mass Characterization},'' \emph{Ultrasound in Medicine
  and Biology}, vol.~40, no.~5, pp. 917--930, 2014. [Online]. Available:
  \url{http://www.sciencedirect.com/science/article/pii/S0301562913011927}
\BIBentrySTDinterwordspacing

\bibitem{W.2012}
W.~Jakubowski, K.~Dobruch-Sobczak, and B.~Migda, ``{Standards of the Polish
  Ultrasound Society – update. Sonomammography examination},'' \emph{J.
  Ultrason}, vol.~50, pp. 245--261, 2012.

\bibitem{Shen2007}
W.~C. Shen, R.~F. Chang, W.~K. Moon, Y.~H. Chou, and C.~S. Huang, ``{Breast
  Ultrasound Computer-Aided Diagnosis Using BI-RADS Features},'' \emph{Academic
  Radiology}, vol.~14, no.~8, pp. 928--939, 2007.

\bibitem{Alvarenga2012}
A.~V. Alvarenga, A.~F.~C. Infantosi, W.~C.~a. Pereira, and C.~M. Azevedo,
  ``{Assessing the combined performance of texture and morphological parameters
  in distinguishing breast tumors in ultrasound images},'' \emph{Medical
  Physics}, vol.~39, no.~12, pp. 7350--7358, 2012.

\bibitem{Alvarenga2010}
A.~V. Alvarenga, A.~F.~C. Infantosi, W.~C.~A. Pereira, and C.~M. Azevedo,
  ``{Assessing the performance of morphological parameters in distinguishing
  breast tumors on ultrasound images},'' \emph{Medical Engineering and
  Physics}, vol.~32, no.~1, pp. 49--56, 2010.

\bibitem{Chou2001}
Y.~H. Chou, C.~M. Tiu, G.~S. Hung, S.~C. Wu, T.~Y. Chang, and H.~K. Chiang,
  ``{Stepwise logistic regression analysis of tumor contour features for breast
  ultrasound diagnosis.}'' \emph{Ultrasound in medicine {\&} biology}, vol.~27,
  no.~11, pp. 1493--1498, 2001.

\bibitem{Chang2005}
R.~F. Chang, W.~J. Wu, W.~K. Moon, and D.~R. Chen, ``{Automatic ultrasound
  segmentation and morphology based diagnosis of solid breast tumors},''
  \emph{Breast Cancer Research and Treatment}, vol.~89, no.~2, pp. 179--185,
  2005.

\bibitem{Joo2004}
S.~Joo, Y.~S. Yang, W.~K. Moon, and H.~C. Kim, ``{Computer-aided diagnosis of
  solid breast nodules: Use of an artificial neural network based on multiple
  sonographic features},'' \emph{IEEE Transactions on Medical Imaging},
  vol.~23, no.~10, pp. 1292--1300, 2004.

\bibitem{Drukker2004a}
K.~Drukker, M.~L. Giger, C.~J. Vyborny, and E.~B. Mendelson, ``{Computerized
  Detection and Classification of Cancer on Breast Ultrasound},''
  \emph{Academic Radiology}, vol.~11, no.~5, pp. 526--535, 2004.

\bibitem{Kim2002}
K.~G. Kim, J.~H. Kim, and B.~G. Min, ``{Classification of malignant and benign
  tumors using boundary characteristics in breast ultrasonograms.}''
  \emph{Journal of digital imaging : the official journal of the Society for
  Computer Applications in Radiology}, vol. 15 Suppl 1, no. March, pp.
  224--227, 2002.

\bibitem{Drukker2005}
K.~Drukker, M.~L. Giger, and C.~E. Metz, ``{Robustness of computerized lesion
  detection and classification scheme across different breast US platforms.}''
  \emph{Radiology}, vol. 237, no.~3, pp. 834--840, 2005.

\bibitem{Drukker2002}
K.~Drukker, M.~L. Giger, K.~Horsch, M.~A. Kupinski, C.~J. Vyborny, and E.~B.
  Mendelson, ``{Computerized lesion detection on breast ultrasound},''
  \emph{Medical Physics}, vol.~29, no.~7, pp. 1438--1446, 2002.

\bibitem{Horsch2002a}
K.~Horsch, M.~L. Giger, L.~a. Venta, and C.~J. Vyborny, ``{Computerized
  diagnosis of breast lesions on ultrasound.}'' \emph{Medical physics},
  vol.~29, no.~2, pp. 157--164, 2002.

\bibitem{Chen2003}
C.-M. Chen, Y.-H. Chou, K.-C. Han, G.-S. Hung, C.-M. Tiu, H.-J. Chiou, and
  S.-Y. Chiou, ``{Breast lesions on sonograms: computer-aided diagnosis with
  nearly setting-independent features and artificial neural networks.}''
  \emph{Radiology}, vol. 226, no.~2, pp. 504--514, 2003.

\bibitem{Chen2004}
S.-C. Chen, Y.-C. Cheung, C.-H. Su, M.-F. Chen, T.-L. Hwang, and S.~Hsueh,
  ``{Analysis of sonographic features for the differentiation of benign and
  malignant breast tumors of different sizes.}'' \emph{Ultrasound in obstetrics
  {\&} gynecology : the official journal of the International Society of
  Ultrasound in Obstetrics and Gynecology}, vol.~23, no.~2, pp. 188--193, 2004.

\bibitem{Su2011}
Y.~Su, Y.~Wang, J.~Jiao, and Y.~Guo, ``{Automatic detection and classification
  of breast tumors in ultrasonic images using texture and morphological
  features.}'' \emph{The open medical informatics journal}, vol.~5, no. Suppl
  1, pp. 26--37, 2011.

\bibitem{Minavathi2012}
M.~S. Minavathi and M.~S. Dinesh, ``{Classification of Mass in Breast
  Ultrasound Images using Image Processing Techniques},'' \emph{International
  Journal of Computer Applications}, vol.~42, no.~10, pp. 29--36, 2012.

\bibitem{Chan2001}
T.~F. Chan and L.~A. Vese, ``{Active contours without edges},'' \emph{IEEE
  Transactions on Image Processing}, vol.~10, no.~2, pp. 266--277, 2001.

\bibitem{daoud2016fusion}
M.~I. Daoud, T.~M. Bdair, M.~Al-Najar, and R.~Alazrai, ``{A Fusion-Based
  Approach for Breast Ultrasound Image Classification Using Multiple-ROI
  Texture and Morphological Analyses},'' \emph{Computational and Mathematical
  Methods in Medicine}, vol. 2016, 2016.

\bibitem{Fawcett2006}
T.~Fawcett, ``{An introduction to ROC analysis},'' \emph{Pattern Recognition
  Letters}, vol.~27, no.~8, pp. 861--874, 2006.

\end{thebibliography}

\end{document}